\documentclass{article}

\usepackage[preprint]{neurips_2023}

\usepackage[utf8]{inputenc} 
\usepackage[T1]{fontenc}    
\usepackage{hyperref}       
\usepackage{url}            
\usepackage{booktabs}       
\usepackage{amsfonts}       
\usepackage{nicefrac}       
\usepackage{microtype}      
\usepackage{xcolor}         
\usepackage{outlines}
\usepackage{amsmath}
\usepackage{dirtytalk}
\usepackage{graphicx}
\usepackage[linesnumbered,ruled]{algorithm2e}
\usepackage{subcaption}
\usepackage{natbib}

\newcommand*{\argmax}{\operatornamewithlimits{argmax}\limits}

\title{Bayesian Optimization of Bilevel Problems}

\author{
  Omer Ekmekcioglu\\
  Warwick Business School,\\
  University of Warwick\\
  \texttt{omer.ekmekcioglu@warwick.ac.uk} \\
  \And
  Nursen Aydin \\
  Warwick Business School,\\
  University of Warwick\\
  \texttt{nursen.aydin@wbs.ac.uk} \\
  \And
  Juergen Branke \\
  Warwick Business School,\\
  University of Warwick\\
  \texttt{Juergen.Branke@wbs.ac.uk} 
}

\begin{document}

\maketitle

\begin{abstract}
Bilevel optimization, a hierarchical mathematical framework where one optimization problem is nested within another, has emerged as a powerful tool for modeling complex decision-making processes in various fields such as economics, engineering, and machine learning. This paper focuses on bilevel optimization where both upper-level and lower-level functions are black boxes and expensive to evaluate. We propose a Bayesian Optimization framework 
that models the upper and lower-level functions as Gaussian processes over the combined space of upper and lower-level decisions, allowing us to exploit knowledge transfer between different sub-problems. Additionally, we propose a novel acquisition function for this model. Our experimental results demonstrate that the proposed algorithm is highly sample-efficient and outperforms existing methods in finding high-quality solutions.
\end{abstract}

\section{Introduction}
Bilevel optimization involves nested optimization problems where the upper-level optimization problem depends on the solution of the lower-level problem. This framework is commonly used to model hierarchical decision-making problems involving interactions between a leader and a follower and is at the core of a wide range of applications \citep{bil_rev_apps}, including self-tuning neural networks \citep{mackay2019self},\citep{stn_2}, network inspection against adversarial attacks \citep{Network_Inspection_Bilevel}, inverse optimal control \citep{bleaq_invopt_ex}, revenue management \citep{bilevel_pricing, Network_pricing_bilevel}, facility location optimization \citep{bil_facility, knap_interdiction} and vehicle routing \citep{bil_vehicle}.

The nested structure of bilevel problems makes them challenging to optimize. Most solution techniques transform the problem into a single optimization problem using lower-level value functions, quasi-variational inequality functions, KKT conditions \citep{bilevel_book}, and tight relaxations \citep{bil_pes_solv}. Recent work focuses on using descent methods to solve bilevel problems \citep{Bilevel_solver1}, with some methods offering theoretical guarantees under specific assumptions \citep{Bilevel_solver2}. However, these algorithms typically assume that functions and gradients are easy to evaluate, which may not always be the case. For example, modeling European gas markets as a bilevel problem with chance constraints requires these constraints to be modeled as a black-box system \citep{black_box_constraint}. Other applications include the design of graph neural networks against adversarial attacks \citep{BO_attack_old}, differential privacy problems where the leader does not know the followers' plans \citep{NN_Priv} and decision-dependent uncertainty problems \citep{bilevel_rev_uncertainty}. These problems motivate our study of bilevel optimization problems where both the upper and lower objective functions are black-box. 

As generalized solution methods, surrogate-assisted metaheuristics have become popular in recent years \citep{metabilevel_rev}. These algorithms treat the upper-level optimization problem as a black-box function and solve the lower level using gradient-based methods in every iteration. \cite{6900529} improve this approach by refining the stopping criteria to reduce computation times. \cite{bleaq} propose using quadratic approximations at the lower level to further enhance computational performance.  Other studies based on evolutionary computation aim to learn different upper-level responses in parallel to utilize transfer learning in the search for the optimal lower-level responses \citep{evo_parallel}. However, the high number of function evaluations and the required knowledge about the lower-level function properties in these algorithms make them unsuitable for expensive black-box optimization problems. Therefore, \cite{6557606} use evolutionary algorithms to solve both upper and lower-level functions, where the lower level is solved for each attempted setting of the upper-level decision variable. This approach, however, has two significant drawbacks: solving the lower level in every iteration is highly sample-inefficient, and the quality of the upper-level solution depends heavily on the accuracy of the lower-level estimates at each iteration.

In recent years, there has been a growing interest in the field of Bayesian Optimization (BO) due to its sample efficiency in solving expensive black-box optimization problems. It has been successfully applied to a wide variety of domains such as hyper-parameter tuning \citep{jin2021autolrs}, robotics \citep{BO_gait}, quantum circuit design \citep{NEURIPS2023_3adb85a3}, experimental design \citep{NEUR_expdes} and drug discovery \citep{Colliandre_Muller_2023, localBO}. These fields often require the optimization of black-box systems, where evaluating the underlying function is expensive, and gradient information is limited. The BO framework addresses these challenges by focusing on inputs with the highest potential for improvement, cleverly selecting which inputs to test, and maximizing efficiency in finding the best configuration.

\subsection{Related Work}
BO has been used as a surrogate-assisted solution method for bilevel problems. One of the earlier studies in the literature models the upper level as a BO problem and solves the lower level function to optimality using Sequential Least Squares Quadratic Programming (SLSQP), under the assumption that the lower-level function is not a black box  \citep{Kieffer_Danoy_Bouvry_Nagih_2017}. This BO method achieves better performance than its competitor BLEAQ which relies on evolutionary algorithms \citep{bleaq}. Generalizing this framework, \cite{8477714} show how using BO in both levels improves the solution quality and sample efficiency compared to evolutionary algorithms. Furthermore, this approach is theoretically studied in the context of hyperparameter tuning of neural networks where the lower level is solved using backpropagation to find the network weights and the upper level is solved using BO to find the right network hyperparameters \citep{bil_converge}. Improving this baseline approach, \cite{Vedatd_new} propose modeling the upper-level Gaussian Process (GP) as a joint function of both upper- and lower-level decision variables. For a given upper-level action, optimal lower-level responses are obtained using SLSQP which assumes that the lower level is not a black box function.

The BO framework for bilevel problems most similar to our work, with black-box functions at both levels, is proposed by \cite{Wang_Singh_Ray_2021}. For each upper-level decision, a separate lower-level optimization problem is solved from scratch using BO. However, this approach has two limitations: efficiency and accuracy. The upper-level surrogate works with noisy observations as the lower level is not guaranteed to find the optimal response given a very restricted budget. Additionally, the nested structure of the problem results in a high overall budget requirement, even when the lower-level evaluation budget is minimal. 

\subsection{Contributions}
The literature contains only a few studies that apply BO to bilevel optimization problems. Furthermore, most of these studies focus on a semi-black-box scenario, where only one level is a black-box function. In contrast, the other level is known and solvable using standard gradient-based optimization techniques, such as hyperparameter tuning in neural networks \citep{bil_converge}, or the application of surrogate-assisted optimization models where the lower level is solved using SLSQP or other off-the-shelf solvers \citep{BO_inverse_opt}.

In some applications, both levels are black-box functions. Existing approaches capable of handling black-box functions at both levels are computationally expensive, as they solve the lower level to a \say{decent} optimality independently and from scratch for every considered upper-level decision \citep{8477714, Wang_Singh_Ray_2021}. While these methods are more efficient than their evolutionary counterparts, they rely on a generous budget for the lower level, limiting their scalability.  Moreover, inaccurate lower-level solutions can adversely impact the algorithm's overall performance. As we will demonstrate, the algorithm proposed in our paper effectively addresses both challenges.

The contributions of this paper are as follows.
\begin{outline}[enumerate]
    \1 We propose a novel framework for applying BO to bilevel optimization problems by modeling each level as a GP over both upper and lower-level decision variables, thereby allowing information collection and knowledge transfer across sub-problems.
    \1 We propose novel acquisition functions inspired by multi-task learning to efficiently sample the decision space. 
    \1 We empirically demonstrate the efficiency and robustness of the proposed approach by comparing it to alternative methods from the literature. 
\end{outline}

In the following sections, we define the problem setting and basic notation that we will use for our algorithm.

\section{Background}
\subsection{Bilevel Optimization Formulation}
The general bilevel optimization problem can be formulated as follows \citep{7942105}:

\begin{align*}
& \max_{x_u \in X_U, x_l \in X_L}{ } F\left(x_u, x_l\right) \\
& \text { s.t. } \\
& \quad x_l \in \underset{z \in X_L}{\operatorname{argmax}}\left\{f\left(x_u, z\right): g_j\left(x_u, z\right) \leq 0, j=1, \ldots, J\right\} \\
& \quad G_k\left(x_u, x_l\right) \leq 0, k=1, \ldots, K
\end{align*}

where $F$ and $f$ represent the upper-level and lower-level functions, and $G_k$ and $g_j$ denote the upper-level and lower-level constraints, respectively. The goal is to find the optimal upper-level action and lower-level response pair $(x_u,x_l)$. The upper-level problem is constrained by the optimal solution of the lower-level problem for any upper-level decision variable.
In this paper, we focus on bilevel optimization problems with only box constraints on both types of decision variables. Therefore, the problem simplifies to

\begin{align*}
& \max_{x_u \in X_U, x_l \in X_L}{ } F\left(x_u, x_l\right) \\
& \text { s.t. } \\
& \quad x_l \in \underset{z \in X_L}{\operatorname{argmax}}f\left(x_u, z \right) \\
\end{align*}

where, $X_L$ and $X_U$ denote the box constraints for the lower and upper-level decision variables, respectively. In this hierarchical problem,  the leader selects an upper-level decision $x_u$, enabling the follower to solve the lower-level optimization problem and determine $x_l$. The resulting pair $(x_u,x_l)$ is then used to calculate the objective function for each agent.

\subsection{Bayesian Optimization}
Let $f:\mathcal{X} \rightarrow \mathbb{R}$ be a black-box function with no information about the function or its derivatives. We can evaluate the function by querying an input point $x \in \mathcal{X} \subset \mathbb{R}^d$ to obtain the observation $y \in \mathbb{R}, y = f(x)$. The goal is to solve the black-box optimization problem $\max_{x \in \mathcal{X}} f(x)$. To achieve this, the black-box function is modeled by a surrogate model, commonly a GP. GPs are characterized by a mean function, $\mu(x)$, and a kernel (covariance function), $k(x,x')$. Using GPs as surrogate models offers a few advantages, such as a closed-form solution for the posterior.

Let $\mathcal{D}^n = (\mathbf{X},\mathbf{y})$ represent the data collected in $n$ observations, we can compute the posterior mean and covariance of a point using:
\begin{align*}
\mu^n(x) & = k(x, \mathbf{X}) \left[ K(\mathbf{X}, \mathbf{X}) + \epsilon^2 I \right]^{-1} \mathbf{y}, \\
k^n(x,x') & = k(x, x') - k(x, \mathbf{X}) \left[ K(\mathbf{X}, \mathbf{X}) + \epsilon^2 I \right]^{-1} k(\mathbf{X}, x'),
\end{align*}

where $\epsilon^2$ denotes the observation noise and $K$ is the Gram matrix. BO is concerned with identifying the next candidate to sample in each iteration, the point that is expected to yield the largest information gain. This is achieved through the optimization of an acquisition function. The general BO framework is summarized in Algorithm \ref{AlgoBOVanilla}. In each iteration $n$, given the dataset, $\mathcal{D}^{n}$, the surrogate GP is fitted to this data, and the next point to sample is determined as $x^{n+1} = \arg\max_{x} \alpha^n(x)$ (Line 5), with $\alpha^n(x)$ being the acquisition function. 
Once the point has been sampled (Line 6), the new data is added to the dataset (Line 7), and the GP model is re-fitted (Line 8). This procedure continues until the sampling budget is exhausted or another termination criterion is met. Upon termination, the best solution is typically chosen in one of two ways (Line 10): either by selecting the best-observed solution or by selecting the maximizer of the surrogate model's posterior mean. This distinction also plays a significant role in constructing acquisition functions such as Expected Improvement (EI) and Knowledge Gradient (KG). 

\begin{algorithm}
    Initialize $\mathcal{D}^{0}$ using Sobol sampling\\ 
    Fit a GP to the initial dataset $\mathcal{D}^{0}$ to initialize $\mathcal{GP}$.\\
    \For{$n \leftarrow 0$ \KwTo $N$}{
    \# Optimize the acquisition function:\\
    $x^{n+1} = \argmax_{x} \alpha^n(x)$\\
    Evaluate: $y^{n+1} = f(x^{n+1})$ \\
    Update: $\mathcal{D}^{n+1} \leftarrow \mathcal{D}^{n} \cup \{ x^{n+1}, y^{n+1} \}$  \\
    Update: $\mathcal{GP}$\\
    } 
    \Return  $\argmax _{x \in \mathcal{X}} \mu^n(x)$ or $\{ x_i \mid i = \arg\max_{j} \{ y_j \mid (x_j, y_j) \in D^{N+1} \} \}$
    \caption{Bayesian Optimization}
    \label{AlgoBOVanilla}
\end{algorithm}

\subsubsection{Knowledge Gradient}
Knowledge Gradient (KG) is one of the most widely used acquisition functions in the BO literature \citep{frazier_tutorial}. In KG, we consider that the maximizer of the posterior mean will be selected by the decision maker. It employs a one-step Bayes-optimal strategy, where in each iteration, the next sampling point is chosen based on its potential to provide the largest expected improvement in the maximum of the posterior mean:
\begin{align}
    \mathrm{KG}^{n}(x):&= \mathbb{E}_{y_{n+1}}\left[\max _{x^{\prime} \in \mathcal{X}} \mu^{n+1}\left(x^{\prime}\right) \mid x_{n+1} = x\right]-\max _{x^{\prime \prime} \in \mathcal{X}} \mu^n\left(x^{\prime \prime}\right).
\end{align}

Due to the intractable expectation term, multiple methods have been proposed for approximation over the years such as the discrete KG \citep{KG_original},  parallel KG \citep{qKG}, hybrid KG \citep{Hybrid_KG_pearce}, One-Shot KG \citep{balandat2020botorch}, and one-shot hybrid KG \citep{Juan_KG}. 

\subsubsection{Multi-Task BO}
Information can be transferred between BO problems if the tasks are correlated by jointly modeling them. The multi-task approach, introduced by \cite{NIPS2013_f33ba15e}, has gained popularity and is related to transferring knowledge from one domain to another in machine learning. They propose using multi-task GPs, which use an intrinsic co-regionalization model to facilitate knowledge transfer between tasks. \cite{Revi_pearce} proposed an acquisition function for querying the next \say{task-action} pair to simultaneously optimize multiple tasks. This acquisition function can also deal with continuous task spaces, which had not been considered in previous studies. Similarly, \cite{NEURIPS2019_7876acb6} introduce a multi-task method by extending the widely used Thompson Sampling (TS) approach to handle multiple tasks. In a multi-task BO problem, given multiple tasks, where \( x \in \mathcal{X} \subseteq \mathbb{R}^d \) represents the tasks and \( a \in \mathcal{A} \subseteq \mathbb{R}^d \) represents the possible actions for each task, along with a black-box objective function \( f: \mathcal{X} \times \mathcal{A} \rightarrow \mathbb{R} \), the goal is to determine a mapping that selects the optimal action for a given task, \( \Phi(x) = \arg\max_a f(x, a) \). 

\section{BILBAO}
In this section, we propose the BILevel BAyesian Optimization algorithm (BILBAO).
The intuition behind our algorithm can be summarized as an alternating optimization method with two steps. The first step involves selecting the upper-level decision variables that maximize the upper-level objective, restricted to the lower-level response map. This response map is derived from the lower-level GP in every iteration. In the second step, we aim to obtain the optimal lower-level response $x_l^* \in \mathbb{R}^{d_l}$, for any candidate upper-level decision $x_u \in \mathbb{R}^{d_u}$, i.e.,  the optimal map $x_l^* = \Phi^*(x_u)$. Previous algorithms find $x_l^*$ by solving the lower-level problem again from scratch for every $x_u$ considered by the upper-level, creating a computational bottleneck as well as making the system highly dependent on the quality of the estimate \citep{Wang_Singh_Ray_2021}. As a remedy, we propose to search for the map $\Phi$ efficiently with a multi-task BO approach. 

In the derivation of the response map $\Phi$, we use the Regional Expected Value of Improvement (REVI) acquisition function. REVI is an extension of KG to multi-task problems proposed by \cite{Revi_pearce} and is used to find the optimal response $x_l^* \in \mathcal{X}_l$ for each upper-level decision $x_u \in \mathcal{X}_u$, i.e., $x_l^* = \Phi^*(x_u)$. Given a probability distribution for the upper-level decision variable, $\mathbb{P}[x_u]$, it can be defined as follows:
{\small{
\begin{align}
\text{REVI}(x_u,x_l) = \int\limits_{x'_u\in\mathcal{X}_u} \mathbb{P}[x'_u]\left(\mathbb{E}_{y^{n+1}}\left[\max _{x_l^{\prime}} \mu^{n+1}(x'_u,x_l^{\prime})-\max _{x_l^{\prime \prime}} \mu^n(x'_u, {x_l^{\prime \prime}}) | (x_u, x_l)^{n+1}=(x_u, x_l)\right]\right)  d x'_u    
\end{align}}}

For computational traceability, REVI discretizes decision spaces and uses Monte Carlo sampling,
{\small{\begin{align}
\operatorname{REVI^{n}}\left(x_u, x_l \right) &= \frac{1}{|X_{UMC}|} \sum_{x_{u_i} \in X_{UMC}} \mathbb{P}[x_{u_i}] \mathbb{E}\left[\max_{x_l^\prime \in X_{LMC}} \mu^{n+1}\left(x_{u_i},x_l^{\prime}\right) - \max_{x_l^{\prime\prime} \in X_{LMC}} \mu^n\left(x_{u_i}, x_l^{\prime\prime}\right) \right]
\label{eq:revi_def0}\\
&:= \frac{1}{|X_{UMC}|} \sum_{x_{u_i} \in X_{UMC}} \mathbb{P}[x_{u_i}] \mathrm{KG}_{x_{u_i}}^{n}(x_u,x_l) \label{eq:revi_def},
\end{align}}}

where  $X_{UMC}$ and $X_{LMC}$  are the Monte Carlo samples and $|X_{UMC}|$ denotes the cardinality of the set. 
$\mathrm{KG}_{x_{u_i}}^{n}(x_u, x_l)$ is the KG for a fixed upper-level decision $x_{u_i}$, i.e., the improvement in the maximum of the posterior mean given the upper-level $x_{u_i}$, if information is collected at location $(x_u, x_l)$. Note that $x_u$ does not have to be in $X_{UMC}$. The probability $\mathbb{P}[x_{u_i}]$  allows the acquisition function to be reweighted by the probability distribution of the upper-level decisions and thus introduces an important link from the upper to the lower level.

We model the upper and lower-level functions using separate GPs, $\mathcal{GP}_U$ for the upper-level function $F$, and $\mathcal{GP}_L$ for the lower-level function $f$. Let $F^{n}(x_u, x_l) \sim \mathcal{GP}_U(\mu^n(x_u, x_l), k^n((x_u, x_l), (x_u', x_l')))$ be a sample path generated from $\mathcal{GP}_U$. We denote the restricted sample path by $F^{n}_\Phi(x_u):= F^{n}(x_u,\Phi(x_u))$.

At a given iteration $n$, we perform two main operations. The first operation is the decision of the upper-level action to be used to evaluate the upper-level objective to update $\mathcal{GP}_U$, which we pick by applying Thompson sampling, i.e., determining the maximizer of the restricted sample path $F^{n}_\Phi(x_u)$. Then we evaluate the upper-level function using the selected upper-level decision and the estimated lower-level response pair, $y^{n+1}_u = F(x^{n+1}_u,\Phi^{n}(x^{n+1}_u))$. We use the new tuple $\{x^{n+1}_u,\Phi^{n}(x^{n+1}_u),y^{n+1}_u\}$ to update $\mathcal{GP}_U$. Following the upper-level update, the algorithm decides on the next query point for the lower level. 
The lower-level acquisition function, denoted as $\alpha^{n}$, is optimized to find the next query point, $(x_u^{n+1},x_l^{n+1}) = \arg\max_{x_u, x_l}\alpha^{n}(x_u, x_l)$. The lower-level GP is then updated with the newly obtained input-output tuple, $\{x^{n+1}_u,x^{n+1}_l,y^{n+1}_l\}$ where $y^{n+1}_l = f(x^{n+1}_u,x^{n+1}_l)$. In our experiments, we tested different acquisition functions at the lower level such as REVI and REVITS, a lightweight alternative to REVI. We describe the general framework in Algorithm \ref{Algo2}.

To calculate the REVI acquisition function in Equation \ref{eq:revi_def}, instead of discretizing the domain of the upper-level decision variables $\mathcal{X}_u$ using Monte Carlo estimates, we focus on specific $x$ values and apply non-uniform weighting across different tasks for calculating the acquisition function. The important step here is to craft the probability distribution across the upper-level decision variables to be used in REVI. We use Thompson sampling to sample the restricted upper-level sample path multiple times to construct a set of important points $\mathcal{X}_{TS} = \{x_u^{*,1}, \dots,x_u^{*,k}\}$, where $x_u^{*,i} = \argmax_{x_u} F^{n,i}_\Phi(x_u)$ for all $i \in \{1,\dots,k\}$. These points are likely candidates to be chosen by the upper-level decision maker. Substituting this probability sample, we obtain the following acquisition function,
\begin{align}
\operatorname{REVI}^{n}(x_u, x_l)= \frac{1}{k} \sum_{x_{u_i} \in\mathcal{X}_{TS}} \mathrm{KG}_{x_{u_i}}^{n}(x_u, x_l). \label{eq:Revi_final}
\end{align}

We optimize Equation \ref{eq:Revi_final} to decide the next query point for the lower level. Intuitively, this acquisition function allows us to improve the map $\Phi(x_u)$ by prioritizing \emph{important} regions determined by the upper level. 

In higher dimensions, REVI can become computationally expensive due to the underlying calculation of discrete KG functions. To balance performance and computational cost, we propose a Thompson sampling-based approach in addition to the REVI acquisition function. We refer to this acquisition function as REVITS, and the corresponding overall algorithm as BILBAO-TS. In this acquisition function, as before we craft a set of \emph{important} upper-level decision variable settings using the upper-level GP, and then apply TS restricted to these upper-level decision variable settings. Specifically, for a given lower-level sample path, $f^{n}(x_u, x_l) \sim \mathcal{GP}_L(\mu^n(x_u, x_l), k^n((x_u, x_l), (x_u', x_l')))$, and a set $\mathcal{X}_{TS}$, we calculate: $(x^{n+1}_u, x^{n+1}_l) = \underset{x_u \in \mathcal{X}_{TS}, x_l}{\arg\max} f^{n}(x_u, x_l)$, to find the next query point for the lower-level function.

\begin{algorithm}
    Initialize $\mathcal{D}^{0}_{l}$ and $\mathcal{D}^{0}_{u}$ using Sobol Sampling\\
    Initialize $\mathcal{GP}_U$ and $\mathcal{GP}_L$ using $\mathcal{D}^{0}_{u}$, $\mathcal{D}^{0}_{l}$.\\
    Initialize $\Phi^{0}$, estimated lower-level response map.\\
    Set $k$: the constant number of interest points.\\
    Given upper-level iterations $N$.\\
    \For{$n \leftarrow 0$ \KwTo $N-1$}{
    \# Thompson Sample $F^{n}_\Phi(x_u)$ \\
    Sample: $F^{n}_\Phi(x_u)$\\
    Find: $x^{n+1}_u = \argmax_{x_u} F^{n}_\Phi(x_u)$\\
    Evaluate: $y^{n+1}_u = F(x^{n+1}_u, \Phi^{n}(x^{n+1}_u))$ \\
    Update: $\mathcal{D}^{n+1}_{u} \leftarrow \mathcal{D}^{n}_{u} \cup \{x^{n+1}_u, \Phi^{n}(x^{n+1}_u), y^{n+1}_u \}$  \\
    Update: $\mathcal{GP}_U$\\
    Thompson Sample:$F^{n+1}_\Phi(x_u)$, $k$ times to generate  $\mathcal{X}_{TS} = \{x_u^{*,1}, \dots,x_u^{*,k}\}$\\
    \# Run REVI using $\mathcal{X}_{TS}$: \\
    $(x^{n+1}_u,x^{n+1}_l) = \argmax \operatorname{REVI}(x_u,x_l | \mathcal{X}_{TS})$ \\
    Evaluate: $y^{n+1}_l = f(x^{n+1}_u,x^{n+1}_l)$\\
    Update: $\mathcal{D}^{n+1}_l \leftarrow \mathcal{D}^{n}_{l} \cup \{ x^{n+1}_u,x^{n+1}_l, y^{n+1}_l \}$\\
    Update: $\mathcal{GP}_L$ and $\Phi$
    } 
    \Return  $\argmax _{x_u} \mu^{N}_u(x_u,\Phi^{N}(x_u))$
    \caption{BILBAO}
    \label{Algo2}
\end{algorithm}

To compute the lower-level response $\Phi$, we discretize the upper-level domain to form a discrete set $X_D$. Then for any $x$, $\Phi(x)$ is calculated by fixing the upper-level decision variables and optimizing the $d_{l}$ dimensional posterior mean function to estimate the lower-level response for each $x_u \in X_D$. Given $\Phi$, a practical approach to calculate Line $8$ is to sample space-filling points for $x_u$, compute the lower-level response $x_l = \Phi(x_u)$ using the current lower level posterior mean  $\mu_l$ and identify the maximum value for $y_u = F(x_u, \Phi(x_u))$. To improve the quality of the discretization set, we include the maximizer of the upper-level posterior mean from all previous iterations.

One key element in the algorithm is how the information is transferred between two levels. To refine the lower-level GP as well as the response map, we re-weigh points that have the potential to maximize the upper-level posterior mean using REVI. This allows the algorithm to make informed decisions at the lower level. The lower-level posterior mean is then used to estimate the lower-level responses for a given upper-level action. This is used to constrain the upper-level posterior mean, making the posterior a function of only upper-level decision variables.

\subsection{Benchmark Algorithm}
\label{subsec:compt}

In this section, we outline a benchmark algorithm adapted from \cite{Wang_Singh_Ray_2021} to highlight the current state-of-the-art in the literature. This algorithm uses only one GP maintained throughout the entire run. It is fitted to a dataset containing only the upper-level decision variables and the estimated response $(x_u, F(x_u,\hat{\Phi}(x_u)))$. In each iteration, a new GP is initialized to estimate the lower-level response by solving the lower-level problem for the specific $x_u$ query point selected by the upper-level. In the ideal case, the lower-level estimates would represent the true optimal lower-level responses, and the collected dataset would be entirely error-free.
However, this may not hold if the lower-level response is difficult to determine within a limited number of BO iterations. In this algorithm, we have to specify how many lower-level iterations $M$ are allowed in each of the $N$ upper-level iterations. Also, we need to specify the initialization budgets $I_u$ and $I_l$ for the upper and lower levels, respectively. 
The total number of evaluations (upper and lower level combined) is then $(I_u+N)(I_l+M+1)$, because for every evaluation of the upper level we require $I_l+M$ evaluations at the lower level. We summarize the benchmark algorithm in Algorithm \ref{Algo_Comp}.

\begin{algorithm}
    Given: Number of upp-level iterations $N$, \\
    Given: Number of lower-level iterations $M$,\\
    Given: Upper-level initialization budget $I_u$\\
    Given: Lower-level initialization budget $I_l$\\
    $\mathcal{D}_u^0 = \emptyset$\\
    $\mathcal{S}_u^0 = \emptyset$\\
    \For{$n \leftarrow 0$ \KwTo $I_u-1$}{
    Initialize $\mathcal{I}_{l}^0$ using Sobol Sampling for a fixed random upper-level point $x^{n}_u$ \\
    Initialize $\mathcal{GP}_L$ using $\mathcal{I}_{l}^0$\\
    \For{$m \leftarrow 0$ \KwTo $M-1$}{
    Find: $x^{m+1}_l = \argmax_{x_l} \alpha_l^{m}(x_l)$\\
    Evaluate: $y^{m+1} = f(x^{n}_u, x^{m+1}_l)$ \\
    Update: $\mathcal{I}_l^{m+1} \leftarrow 
    \mathcal{I}_l^m \cup \{ x^{m+1}_l, y^{m+1}_l \}$ \\
    Update: $\mathcal{GP}_L$\\
    }
    $x_l^i \in \{ x_l^j \mid j = \arg\max_{j} \{ y_l^j \mid (x_l^j, y_l^j) \in \mathcal{I}_l^{M} \} \}$\\
    $y_u^n = F(x_u^n, x_l^i)$\\
    $\mathcal{S}_u^{n+1} = \mathcal{S}_u^n \cup \{x_u^n, y_u^n\}$\\
    }
    $\mathcal{D}_{u}^{0} = \mathcal{S}_u^{k+1}$\\
    Initialize $\mathcal{GP}_U$ using $\mathcal{D}_{u}^{0}$\\ 
    \For{$n \leftarrow 0$ \KwTo $N-1$}{
    \# Optimize the acquisition function:\\
    Find: $x^{n+1}_u = \argmax_{x_u} \alpha_u^{n}(x_u)$\\
    \# Find the optimal response:\\
    Initialize $\mathcal{D}_{l}^0$ using Sobol Sampling and the fixed point $x^{n+1}_u$\\
    Initialize $\mathcal{GP}_L$ using $\mathcal{D}_{l}^0$ \\
    \For{$m \leftarrow 0$ \KwTo $M-1$}{
    \# Optimize the acquisition function:\\
    Find: $x^{m+1}_l = \argmax_{x_l} \alpha_l^{m}(x_l)$\\
    Evaluate: $y^{m+1} = f(x^{n+1}_u, x^{m+1}_l)$ \\
    Update: $\mathcal{D}_l^{m+1} \leftarrow 
    \mathcal{D}_l^m \cup \{ x^{m+1}_l, y^{m+1}_l \}$  \\    
    }
    $x_l^i \in \{ x_l^j \mid j = \arg\max_{j} \{ y_l^j \mid (x_l^j, y_l^j) \in \mathcal{D}_l^{M} \} \}$\\
    Evaluate: $y^{n+1}_u = F(x^{n+1}_u, x^{i}_l)$ \\
    Update: $\mathcal{D}_u^{n+1} \leftarrow \mathcal{D}_u^n \cup \{ x^{n+1}_u, y^{n+1}_u \}$  \\
    Update: $\mathcal{GP}_U$\\
    } 
    \Return $\{ x_u^i \mid i = \arg\max_{j} \{ y_u^j \mid (x_u^j, y_u^j) \in \mathcal{D}^{N} \} \}$
    \caption{Benchmark Algorithm}
    \label{Algo_Comp}
\end{algorithm}

In the algorithm, we denote the upper-level function at iteration $n$ by $\alpha_u^{n}$ and the lower-level acquisition function at iteration $m$ by $\alpha_l^{m}$. In the numerical experiments, we use EI for the acquisition function. Given a fixed $x^{n+1}_u$, the inner loop solves a single-dimensional BO problem at the lower level. In each iteration, a new GP is initialized with a randomly selected set of initial points. The same upper-level candidate solution $x^{n+1}_u$ is used throughout the evaluation of the lower-level function to reduce its dimensionality and make the function depend only on the lower-level decision variables. For notational simplicity, we omit the fixed upper-level decision variable in the lower-level dataset, although it could equivalently be written as $D_l^m = \{ ( x^{n+1}_{u}, x^{m}_{l}), y^m_l \}$, for all $m \in \{1,\dots,M\}$. Therefore, in this algorithm, the lower-level problem is solved in the inner loop, and the upper-level problem is solved in the outer loop. However, this approach has a significant limitation in addition to its computational complexity. The algorithm could get stuck at a sub-optimal solution if the inner loop fails to solve the problem to optimality. This issue is particularly pronounced in high-dimensional, challenging problems. 

To address this issue at least partially, \cite{Wang_Singh_Ray_2021} suggest the option to re-optimize the lower-level response of the best-found upper-level decision variable at the end. However, we did not implement this option here.

\section{Numerical Experiments}
To evaluate the performance of our algorithms, we conduct two tests on synthetic 2D problems and four synthetic 4D problems. For the 2D tests, we construct bilevel problems by combining two-dimensional test functions, assigning the first dimension as the upper-level decision variable and the second dimension as the lower-level decision variable. The test functions are normalized to unit bounds to ensure identical domains. Higher-dimensional tests are performed using functions from the SMD test suite \citep{SMD_Paper}. In all experiments, we have used BoTorch \citep{balandat2020botorch} to fit the GP models and optimize the acquisition functions. For the acquisition functions, we apply the built-in Thompson Sampling and Expected Improvement functions, while implementing the REVI function, which utilizes discrete KG calculations in line with \cite{Revi_pearce}. For the 2D problems, we use 150 points generated by Sobol sampling for discretizing the lower-level decision variable to represent the set $X_{LMC}$ in Equation \ref{eq:revi_def0}. For 4D problems, we increase the number of discrete points to 250. We normalize both the upper-level and lower-level functions to improve the estimates of GPs. For Thompson sampling, we sample $1024$ points in 2D and $2048$ points in 4D problems using Sobol sampling. 

To calculate the lower-level responses, $x_l = \Phi^{n}(x_u)$, we fix the upper-level decision variables and optimize the posterior mean of the lower-level GP, to obtain the corresponding lower-level response for any upper-level action. Specifically, for a given point $x_u$, we solve the optimization problem $\mathbf{x}_l^*(x_u) = \arg\max_{\mathbf{x}_l} \ \mu^n(\mathbf{x}_u, \mathbf{x}_l)$ using L-BFGS-B with $30$ restarts. 

We report the optimality gap at each iteration by assessing the algorithm's performance as if it were terminated at that point. Specifically, at each iteration $n$, we calculate $x_u^n = \arg\max_{x_u} \mu(x_u,\Phi^n(x_u))$ and report the value of $|F(x_u^n, \Phi^*(x_u^n) - F(x_u^*,\Phi^*(x_u^*))|$, where $\phi^*$ denotes the optimal lower level response. To measure the quality of the response map across the entire domain, we define an additional metric, the action gap, as the absolute difference between the function values of the true optimal action and the estimated action. For this, we randomly sample a set $X_e$ of $300$ upper-level actions and calculate their corresponding optimal lower-level responses. At each iteration, we evaluate the function values using the lower-level estimates and compare them with the optimal lower-level responses, so

\begin{equation}
   \text{action gap }= \sum_{x_u\in X_e}|f(x_u,\Phi(x_u)) - f(x_u,\Phi^*(x_u))|.
\end{equation}

Because the compared algorithms use different numbers of evaluations per iteration, we plot performance vs.\ function evaluations, assuming upper and lower-level functions are equally expensive to evaluate.

\subsection{2D Tests}
In the initial set of tests, we select the upper-level function to be a two-dimensional Six-Hump Camel function and the lower-level function to be the Branin function \citep{molga2005test}. The total evaluation budget for BILBAO is set to $180$, distributing it as $10$ for initializing each GP, $80$ for upper-level evaluations, and $80$ for lower-level evaluations. We run the tests $10$ times and report the mean of the observed values. In the second set of tests, following the same setup principles, we used the Dixon-Price function for the upper level and the Branin function for the lower level. In both tests, we select $10$ upper-level decision points to construct the set $X_{TS}$, used in BILBAO and BILBAO-TS.

We denote the benchmark algorithm discussed in Section \ref{subsec:compt} by \say{Benchmark}, and \say{Benchmark 2}. At both the upper and lower levels, we adopt EI as the acquisition function. However, as it is not clear how many lower-level evaluations should be used for each upper-level iteration, different lower-level budgets are used for different benchmarks. In both benchmarks, the total evaluation budget is set to $180$. For \say{Benchmark}, we allocate $3$ evaluations per GP initialization, $20$ for upper-level evaluations, and $4$ for lower-level evaluations. For \say{Benchmark 2}, we use $3$ evaluations per initialization, $27$ for upper-level evaluations, and $2$ for lower-level evaluations. In all of the algorithms, Sobol sampling is used to select random initial points.

\begin{figure}
\centering
\begin{subfigure}{0.45\textwidth}
    \includegraphics[width=\textwidth]{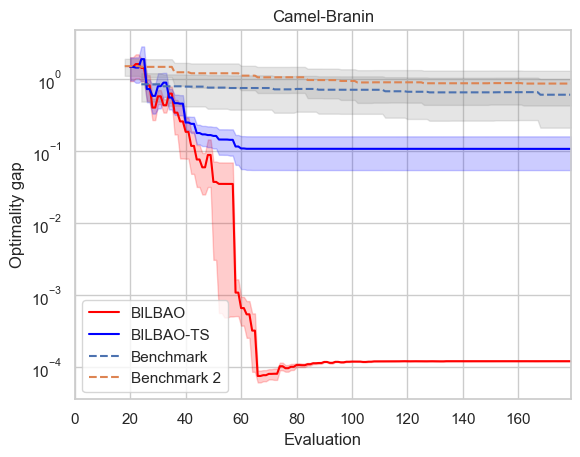}
    \caption{Optimality gap}
    \label{fig:camel_opt}
\end{subfigure}
\hfill
\begin{subfigure}{0.45\textwidth}
    \includegraphics[width=\textwidth]{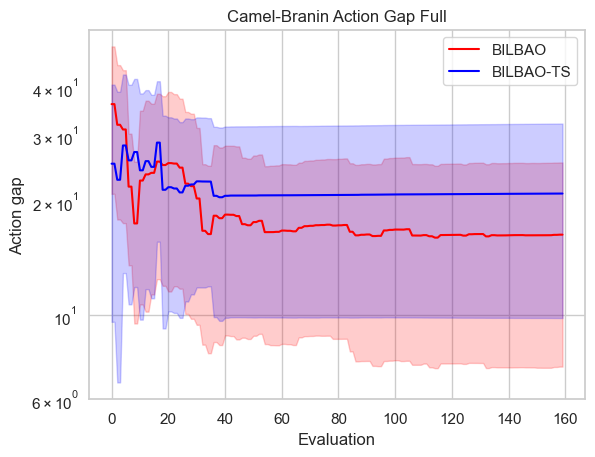}
    \caption{Action gap}
    \label{fig:camel_map}
\end{subfigure}
\caption{Camel-Branin Tests}
\label{Camel Tests}
\end{figure}

\begin{figure}
\centering
\begin{subfigure}{0.45\textwidth}
    \includegraphics[width=\textwidth]{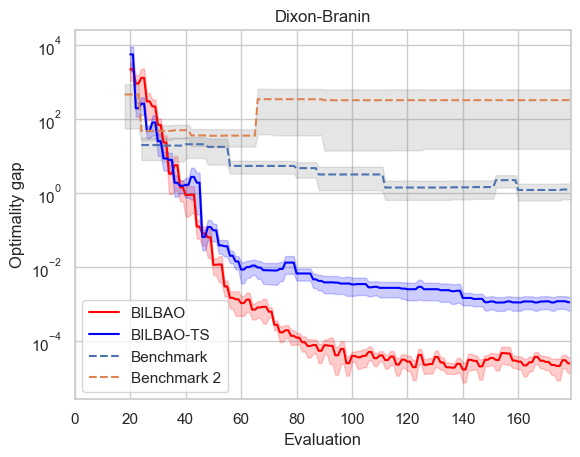}
    \caption{Optimality gap}
    \label{fig:dixon_opt}
\end{subfigure}
\hfill
\begin{subfigure}{0.45\textwidth}
    \includegraphics[width=\textwidth]{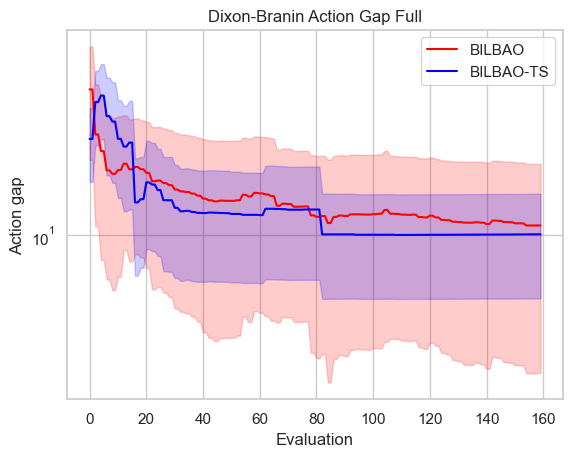}
    \caption{Action gap}
    \label{fig:dixon_act}
\end{subfigure}
\caption{Dixon-Branin Tests}
\label{Dixon Tests}
\end{figure}

We present the optimality gap results in Figures \ref{fig:camel_opt} and \ref{fig:dixon_opt} and the action gaps in Figures \ref{fig:camel_map} and \ref{fig:dixon_act}. The results represent the average of 10 runs, with the shaded regions indicating the standard error across these runs. Note that the action gap metric cannot be reported for the benchmark algorithm, as it recalculates the optimal response from scratch at each iteration. 

In Figures \ref{fig:camel_opt} and \ref{fig:dixon_opt}, we observe that the optimality gap for our algorithm declines faster than the benchmark algorithms. We also observe significant variability in the performance of the benchmark algorithm when different evaluation budgets are used at the lower level. If the lower-level algorithm fails to provide accurate solutions, the resulting suggested solutions can be quite poor, as demonstrated in Benchmark 2.

Figures \ref{fig:camel_map} and \ref{fig:dixon_act} display the decrease in action gap over evaluations. The action gap stabilizes at some point for both algorithms as they only focus on the points that improve the region of interest. Rather than learning the entire map, the algorithms concentrate on a fraction of the domain to learn the lower-level responses for specific upper-level decision variables.

When comparing the performance of BILBAO and BILBAO-TS, it is evident that BILBAO achieves superior results. However, this improvement comes at a somewhat higher computational cost because  REVI is computationally more expensive than TS. Both methods perform similarly in the initial iterations, but BILBAO demonstrates continued improvement in solution quality due to its more advanced approach to selecting the next candidate point. REVI freely identifies a candidate point that maximizes the increase in the average posterior mean across the interest region. In contrast, REVITS is constrained to select a point from the upper-level slices used as interest points, limiting its effectiveness over the long term. Also, REVI aims to improve the lower-level response map at all interest points simultaneously, whereas REVITS focuses on improving only one response.

\begin{figure}
\centering
\begin{subfigure}{0.45\textwidth}
    \includegraphics[width=\textwidth]{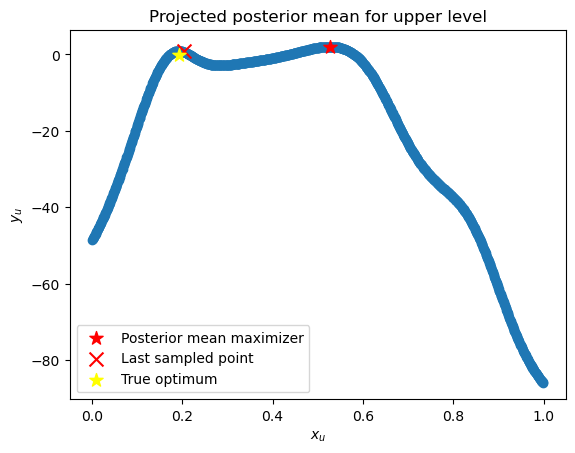}
    \caption{Posterior mean after the first evaluation}
    \label{fig:eval1_proj}
\end{subfigure}
\hfill
\begin{subfigure}{0.45\textwidth}
    \includegraphics[width=\textwidth]{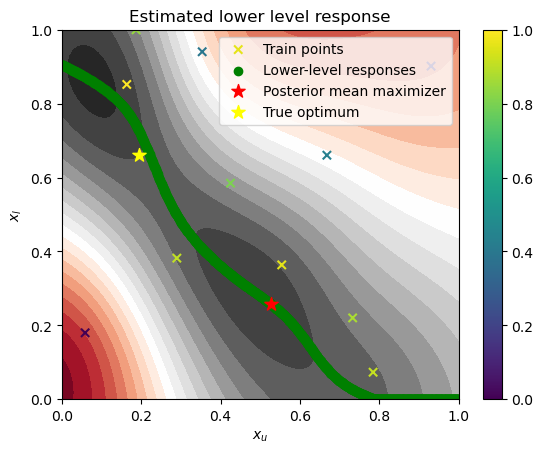}
    \caption{Posterior mean and response map after the first evaluation}
    \label{fig:eval1_map}
\end{subfigure}
\hfill
\begin{subfigure}{0.45\textwidth}
    \includegraphics[width=\textwidth]{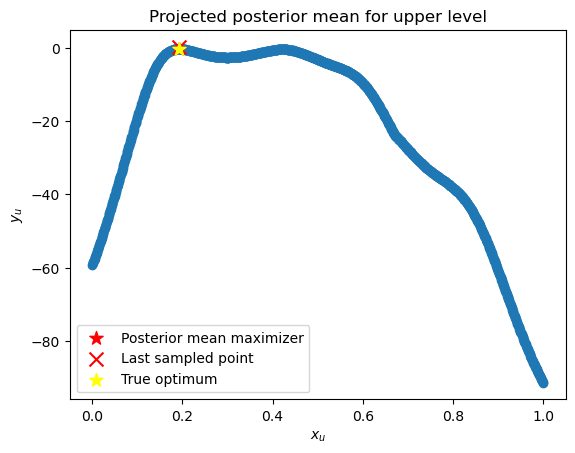}
    \caption{Posterior mean after the 50'th evaluation}
    \label{fig:eval50_proj}
\end{subfigure}
\hfill
\begin{subfigure}{0.45\textwidth}
    \includegraphics[width=\textwidth]{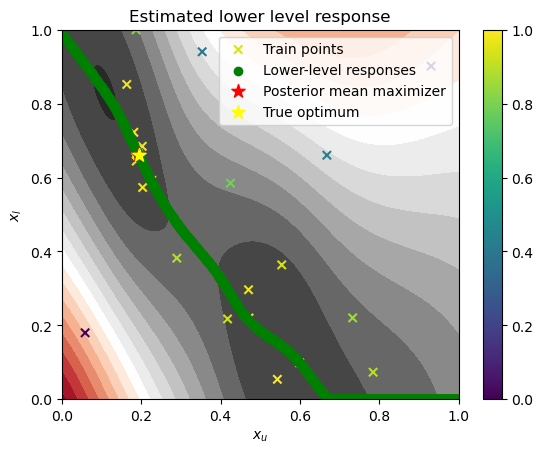}
    \caption{Posterior mean and response map after the 50'th evaluation}
    \label{fig:eval50_map}
\end{subfigure}
\hfill
\begin{subfigure}{0.45\textwidth}
    \includegraphics[width=\textwidth]{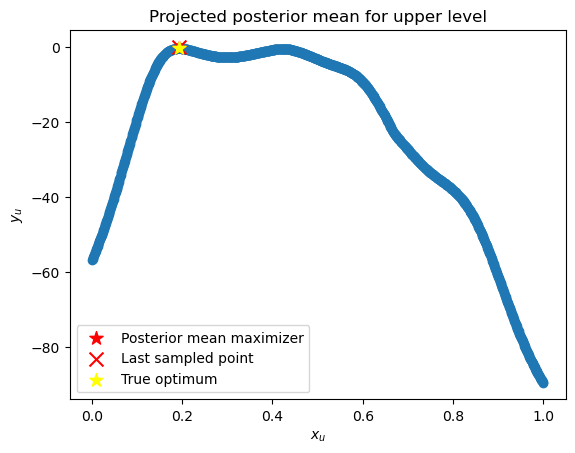}
    \caption{Posterior mean and response map after the 160'th evaluation}
    \label{fig:eval160_proj}
\end{subfigure}
\hfill
\begin{subfigure}{0.45\textwidth}
    \includegraphics[width=\textwidth]{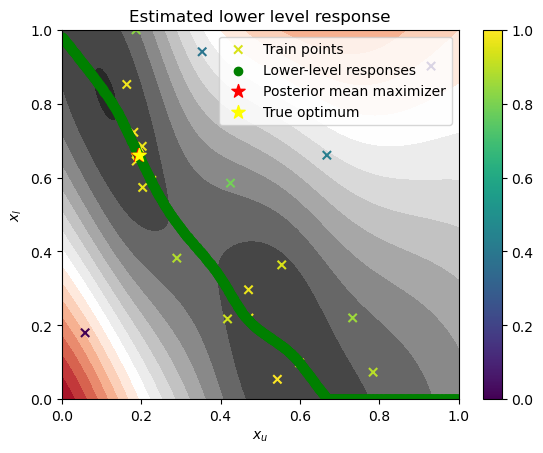}
    \caption{Posterior mean after the 160'th evaluation}
    \label{fig:eval160_map}
\end{subfigure}
\caption{Upper-level restricted posterior mean, $\mu^n(x_u,\Phi^n(x))$, for the Camel-Branin test in different iterations.}
\label{Camel_evolve}
\end{figure}

In Figure \ref{Camel_evolve}, we plot how the lower-level response map, $\Phi^n(x)$, and the restricted posterior mean of the upper-level GP evolves when the REVITS acquisition function has been used. Each subplot represents a snapshot at iterations $1$, $50$ and $160$, corresponding to $\Phi^{1}(x_u)$, $\Phi^{50}(x_u)$, and $\Phi^{160}(x_u)$, respectively. The plots show that the lower-level map learns the best response sample efficiently, enabling the upper-level to sample actions with a well-informed lower-level estimate. Key markers in the plots include the red cross, indicating the last upper-level value sampled, the yellow star denoting the true optimal solution, the red star representing the best action selected in that iteration as the maximizer of the posterior mean, and the green path showing the estimated lower-level responses.

The progression of the lower-level response map in Figure \ref{Camel_evolve} aligns with the optimality gap and action gap results presented in Figure \ref{Camel Tests}. The lower-level responses converge quickly and remain almost identical beyond iteration $50$. Similarly, the true bilevel optimum is identified around iteration $50$.

\subsection{4D Tests}
For the higher-dimensional tests, we use 4D test functions from the SMD benchmark suite \citep{SMD_Paper}. We employ the first four SMD functions and use the first two dimensions as the upper-level decision variables and the remaining two as the lower-level responses. For BILBAO, we set a total evaluation budget of $240$, allocating $20$ for initializing each GP, $100$ for upper-level evaluations, and $100$ for lower-level evaluations. The performance of the benchmark algorithm is assessed under two different budget configurations. In the \say{Benchmark} algorithm, the total evaluation budget is also set to $240$, distributed as $5$ for initializing each GP, $10$ for upper-level evaluations, and $10$ for lower-level evaluations. Similarly, in \say{Benchmark 2}, the total and initialization budgets remain unchanged, but the upper-level budget is adjusted to $17$ and the lower-level budget is reduced to $5$. Across all algorithms, Sobol sampling is used to select random points for initialization.

In the calculation of REVI and REVITS, similar to the 2D tests, we construct the set $X_{TS}$ from $10$ upper-level decision points and use $250$ points to discretize the lower-level action space. 

\begin{figure}[!htb]
\centering
\begin{subfigure}{0.45\textwidth}
    \includegraphics[width=\textwidth]{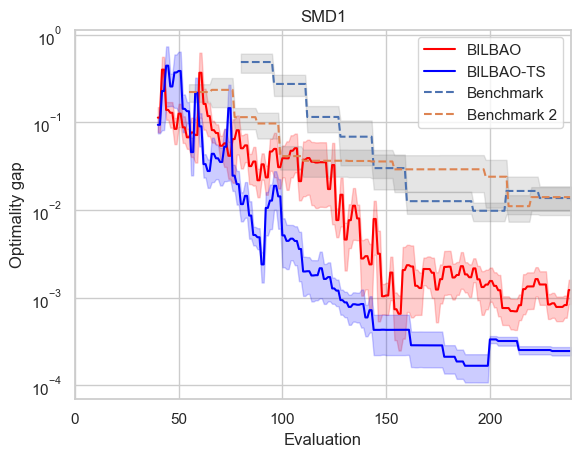}
    \caption{SMD1}
    \label{fig:smd1_log}
\end{subfigure}
\hfill
\begin{subfigure}{0.45\textwidth}
    \includegraphics[width=\textwidth]{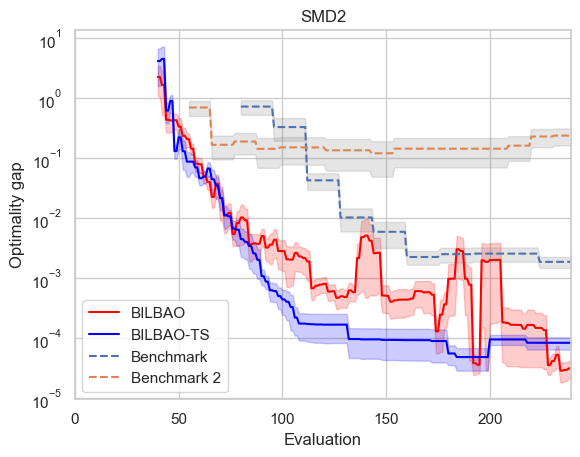}
    \caption{SMD2}
    \label{fig:smd2_log}
\end{subfigure}
\hfill
\begin{subfigure}{0.45\textwidth}
    \includegraphics[width=\textwidth]{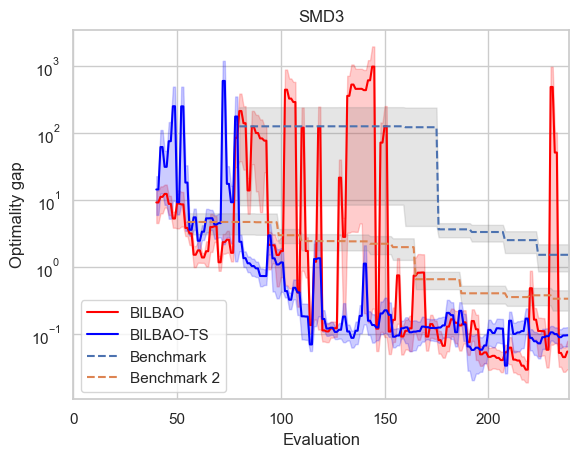}
    \caption{SMD3}
    \label{fig:smd3_log}
\end{subfigure}
\hfill
\begin{subfigure}{0.45\textwidth}
    \includegraphics[width=\textwidth]{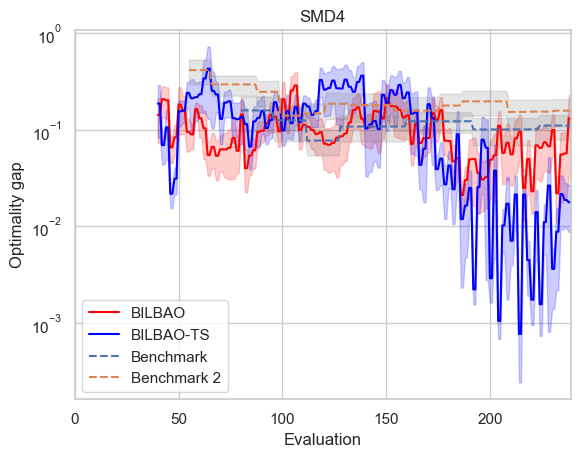}
    \caption{SMD4}
    \label{fig:smd4_log}
\end{subfigure}
\caption{SMD optimality gap logscaled}
\end{figure}

The test results for SMD1 through SMD4 are shown in Figures \ref{fig:smd1_log} to \ref{fig:smd4_log}. For all functions, both BILBAO and BILBAO-TS converge faster than the benchmarks. In Figure \ref{fig:smd4_log}, we observe that the benchmark becomes stuck at a local optimum at the lower level, leading to inaccurate solutions. 

\section{Conclusion}
In this paper, we proposed a new framework for bilevel Bayesian optimization, BILBAO, where both levels are expensive black-box functions. Our approach leverages multi-task Bayesian optimization to learn lower-level responses for promising upper-level decision variables efficiently. We use Thompson sampling on the upper-level GP to explore candidate solutions conditioned on the lower-level response map obtained by the lower-level GP. We show empirically that this framework converges faster and more consistently than state-of-the-art benchmarks.

\newpage
\nocite{*}
\bibliographystyle{agsm}
\bibliography{biblio}
\newpage

\section{Appendix}
\begin{figure}[!htb]
\centering
\begin{subfigure}{0.45\textwidth}
    \includegraphics[width=\textwidth]{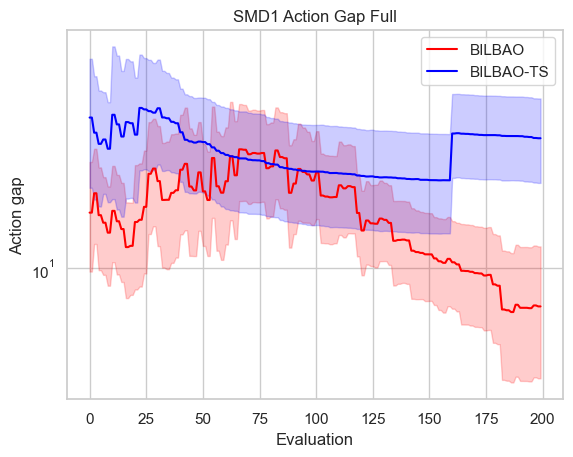}
    \caption{SMD1}
    \label{fig:smd1_act_full}
\end{subfigure}
\hfill
\begin{subfigure}{0.45\textwidth}
    \includegraphics[width=\textwidth]{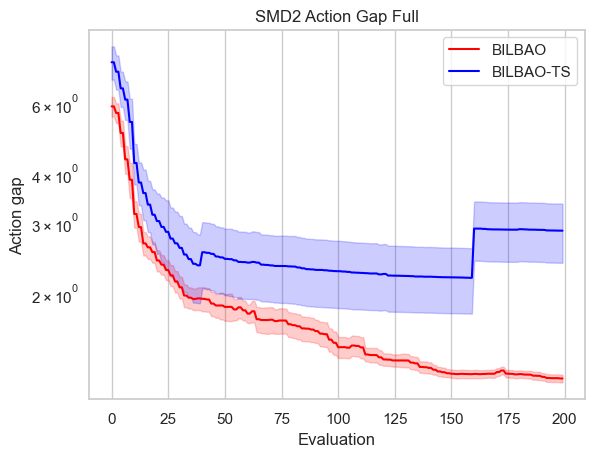}
    \caption{SMD2}
    \label{fig:smd2_act_full}
\end{subfigure}
\hfill
\begin{subfigure}{0.45\textwidth}
    \includegraphics[width=\textwidth]{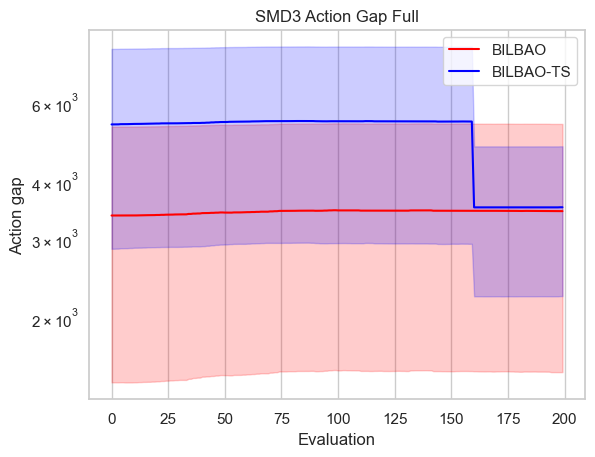}
    \caption{SMD3}
    \label{fig:smd3_act_full}
\end{subfigure}
\hfill
\begin{subfigure}{0.45\textwidth}
    \includegraphics[width=\textwidth]{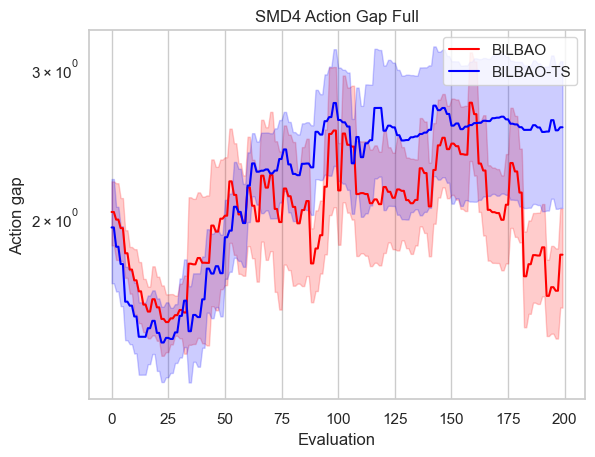}
    \caption{SMD4}
    \label{fig:smd4_act_full}
\end{subfigure}
\caption{Action gaps for the SMD functions}
\label{Fig:ActGapFig}
\end{figure}

\begin{figure}
\centering
\begin{subfigure}{0.45\textwidth}
    \includegraphics[width=\textwidth]{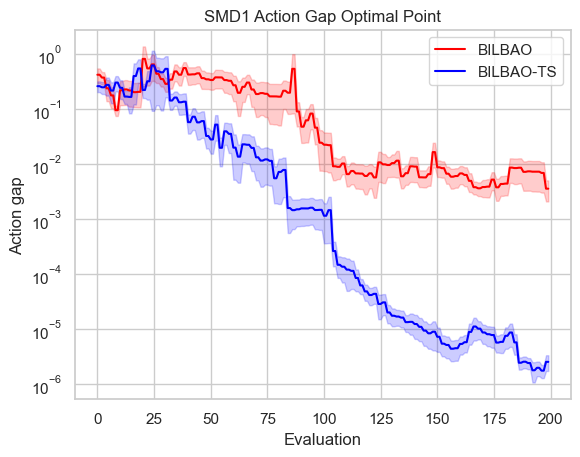}
    \caption{SMD1}
    \label{fig:smd1_act_point}
\end{subfigure}
\hfill
\begin{subfigure}{0.45\textwidth}
    \includegraphics[width=\textwidth]{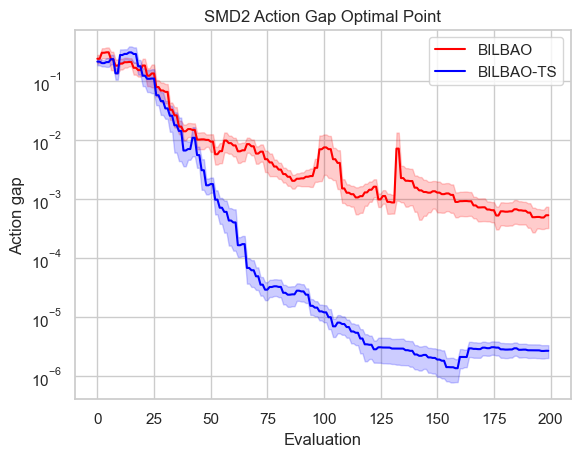}
    \caption{SMD2}
    \label{fig:smd2_act_point}
\end{subfigure}
\hfill
\begin{subfigure}{0.45\textwidth}
    \includegraphics[width=\textwidth]{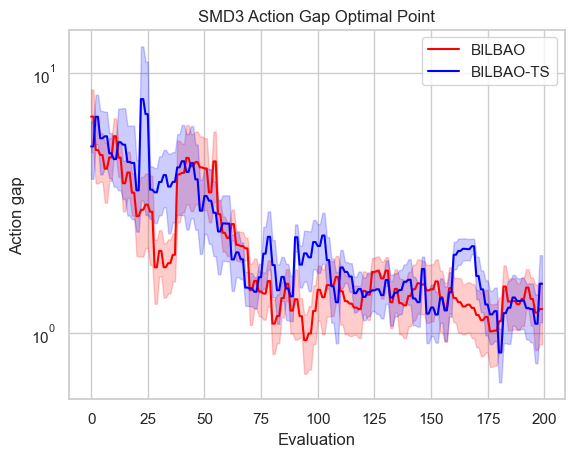}
    \caption{SMD3}
    \label{fig:smd3_act_point}
\end{subfigure}
\hfill
\begin{subfigure}{0.45\textwidth}
    \includegraphics[width=\textwidth]{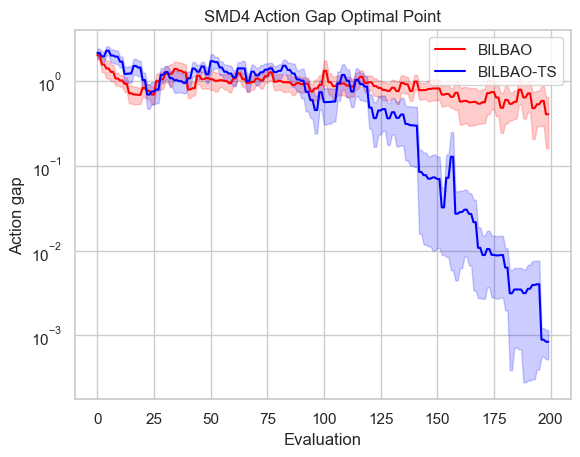}
    \caption{SMD4}
    \label{fig:smd4_act_point}
\end{subfigure}
\caption{Action gaps for the SMD functions only at the optimum}
\label{Fig:ActGapFig_point}
\end{figure}

In Figures \ref{Fig:ActGapFig} and \ref{Fig:ActGapFig_point}, we present the action gaps across the entire upper-level decision domain and the action gap for the optimal bilevel solution, respectively. We present these figures to illustrate the evolution of the complete lower-level response map over iterations, alongside the lower-level response corresponding to the true optimal solution. We highlight our method of leveraging interest regions by demonstrating how our approach identifies key areas of the lower-level map to enhance sampling efficiency without having to learn the entire response map. The plots show the means and the standard errors of $10$ independent runs. 

In Figure \ref{Fig:ActGapFig}, it can be seen that the loss trend does not fully converge. It only decreases when REVI is used in SMD1 and SMD2. This is because the lower-level acquisition functions are designed to sample selectively at points that are informative for the upper level, leaving the majority of the map unsampled. 
If we just look at the action gap at the true optimal upper-level solution, $|F(x_u^*,\Phi^*(x_u^*)-F(x_u^*,\Phi^n(x_u^*)|$ we would expect that the gap keeps reducing as the algorithm hones in onto the optimum, which seems not fully confirmed in 
Figure \ref{Fig:ActGapFig_point}.

\end{document}